%% file: main.tex
\documentclass{bmvc2k}

\usepackage{graphicx}
\usepackage{amsmath}
\usepackage{amsfonts}
\usepackage{booktabs}
\usepackage{adjustbox}
\usepackage{footnotebackref}
\usepackage{multirow}
\usepackage{array}
\usepackage{makecell}
\usepackage{mathtools}
\usepackage{capt-of}
\usepackage{cuted}
\usepackage{xfrac}
\usepackage{microtype}
\usepackage{here}
\usepackage{wrapfig}
\usepackage{xspace}
\usepackage{grffile}
\usepackage{varioref}
\usepackage[capitalize]{cleveref}

\title{A Structure-Guided Diffusion Model for Large-Hole Image Completion}

\addauthor{Daichi Horita}{horita@hal.t.u-tokyo.ac.jp}{1}
\addauthor{Jiaolong Yang}{jiaoyan@microsoft.com}{2}
\addauthor{Dong Chen}{doch@microsoft.com}{2}
\addauthor{Yuki Koyama}{koyama.y@aist.go.jp}{3}
\addauthor{Kiyoharu Aizawa}{aizawa@hal.t.u-tokyo.ac.jp}{1}
\addauthor{Nicu Sebe}{niculae.sebe@unitn.it}{4}

\addinstitution{
 The University of Tokyo
}
\addinstitution{
 Microsoft Research Asia
}
\addinstitution{
 National Institute of Advanced Industrial Science and Technology (AIST)
}
\addinstitution{
 University of Trento
}

\runninghead{Horita et al.}{A Structure-Guided Diffusion Model}

\def\eg{\emph{e.g}\bmvaOneDot}

\def\etal{\emph{et al}\bmvaOneDot}

\makeatletter
\DeclareRobustCommand\onedot{\futurelet\@let@token\@onedot}
\def\@onedot{\ifx\@let@token.\else.\null\fi\xspace}
\def\ie{\emph{i.e}\onedot} 
 
\def\etc{\emph{etc}\onedot}

\def\etal{\emph{et al}\onedot}
\makeatother

\renewcommand{\paragraph}{\vspace{1mm}\noindent\textbf}
\input{floats}

\begin{document}

\maketitle

\figteaser

\begin{abstract}
  \input{abstract}
\end{abstract}

\input{body}

\bibliography{egbib}

\input{supps_body}

\end{document}

%% file: floats.tex
\definecolor{redd}{rgb}{1.0 0.29 0}
\definecolor{blue}{rgb}{0.0, 0.35, 1.0}
\definecolor{greenn}{rgb}{0.011, 0.686, 0.478}
\definecolor{cadmiumgreen}{rgb}{0.0, 0.42, 0.24}
\definecolor{ceruleanblue}{rgb}{0.16, 0.32, 0.75}

\newcommand{\A}[1]{\textcolor{redd}{#1}}
\newcommand{\B}[1]{\textcolor{blue}{#1}}

\newcommand{\figteaser}{
  \begin{figure}[t]
    \centering
    \includegraphics[width=\textwidth]
    {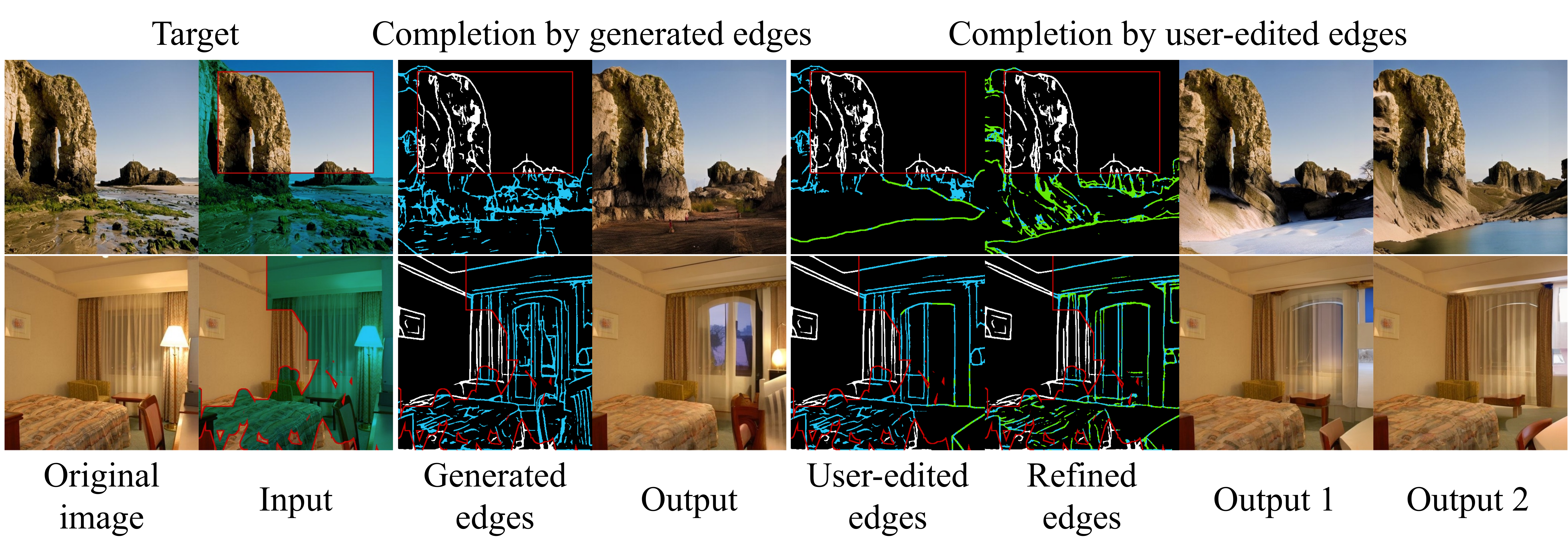}
    \vspace{-0.7cm}
    \caption{
      The SGDM first generates edges within missing regions, indicated by blue.
      Then, it produces textured images using the edges as structural guidance.
      Optionally, the edges can be manually edited, which are then refined by SDEdit \cite{meng2022sdedit} using the SGDM's prior, represented by green.
      The SGDM's stochastic process allows for generating diverse outputs.
    }
    \label{fig:teaser}
  \end{figure}
}

\newcommand{\figindivjointtile}{
  \begin{figure}[t]
    \centering
    \includegraphics[width=\linewidth]{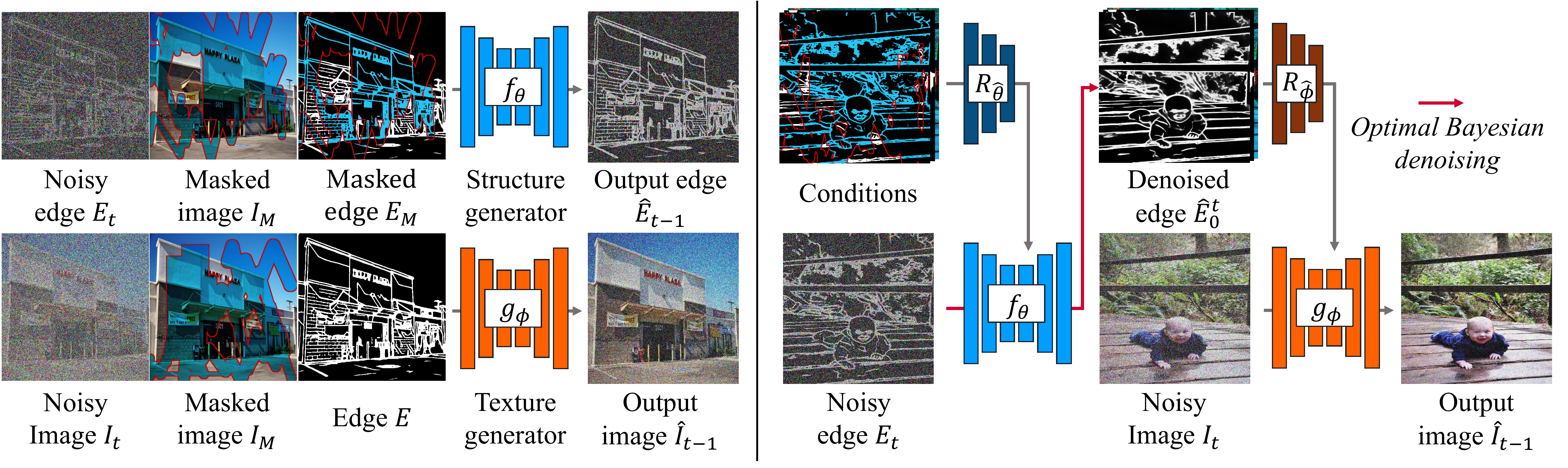}
    \vspace{-0.7cm}
    \caption{
      Overview of our individual training (left) and joint training (right).
    }
    \label{fig:framework}
  \end{figure}
}

\newcommand{\tabablationva}{
  \begin{table}[t]
    \centering
    \footnotesize
    \begin{tabular}{@{}l c c c@{}}
        \toprule
        Places &  & \multicolumn{2}{c}{Large mask} \\
        \cmidrule(lr){3-4}
        Model  & Samples & FID $\downarrow$ & LPIPS $\downarrow$ \\
        \midrule
        (a) Indiv\onedot only & 25M & 32.28 & 0.188 \\
        (b) + Joint w/o OBd & 0.1M & 28.68 & 0.175 \\
        (c) + Joint w OBd & 0.1M & \A{27.47} & \B{0.170} \\
        ~& 1M & \B{27.81} & \A{0.168}  \\
        \bottomrule
    \end{tabular}
    \vspace{0.3cm}
    \caption{
      Result of the ablation study on Places with large masks.
      Indiv\onedot and Joint indicate individual and joint training settings, respectively.
      OBd represents optimal Bayesian denoising.
    }
    \label{tab:ablation_study}
  \end{table}
}

\newcommand{\tabfaces}{
  \begin{table*}[t]
    \centering
    \renewcommand\arraystretch{1.1}
    \setlength\tabcolsep{2pt}
\resizebox{\linewidth}{!}{

\begin{tabular}{@{}l c c c c c c c c c c c c c @{}}
\toprule
\multirow{3}{*}{Method} & \multirow{3}{*}{Modeling} & \multicolumn{6}{c}{Places (512 $\times$ 512)} & \multicolumn{6}{c}{CelebA-HQ (512 $\times$ 512)} \\
				\cmidrule(lr){3-8} \cmidrule(lr){9-14}
				~ & ~ & \multicolumn{3}{c}{Small mask} & \multicolumn{3}{c}{Large mask} & \multicolumn{3}{c}{Small mask} & \multicolumn{3}{c}{Large mask} \\
				\cmidrule(lr){3-5} \cmidrule(lr){6-8} \cmidrule(lr){9-11} \cmidrule(lr){12-14}
~ & ~ & FID$\downarrow$ & P-IDS$\uparrow$ & U-IDS$\uparrow$ & FID$\downarrow$& P-IDS$\uparrow$ & U-IDS$\uparrow$ & FID$\downarrow$ & P-IDS$\uparrow$ & U-IDS$\uparrow$ & FID$\downarrow$ & P-IDS$\uparrow$ & U-IDS$\uparrow$ \\
\midrule
SGDM (ours) & DM & \A{3.85} & \B{25.54} & \A{38.53} & \A{6.96} & \B{18.12} & \B{31.78} & \A{2.58} & \A{22.01} & \A{33.56} & \A{4.72} & \A{13.99} & \A{24.97} \\
\midrule
Stable Diffusion~\cite{sd21inpaint}       & DM  & 5.36 & 16.32 & 32.05 & 7.21 & 15.34 & 30.80 & -    &  -    &  -    &  -    &  -   &  -  \\
LDM~\cite{ldm} & DM  & 5.64 & 13.42 & 30.66 & 8.74 & 11.7 & 27.00 & -    &  -    &  -    &  -    &  -   &  -  \\
MAT~\cite{li2022mat}             & GAN & 4.10 & \A{25.56} & \B{37.73} & \B{7.11} & \A{18.40} & \A{32.46} & \B{2.81} & \B{19.24} & \B{31.33} & \B{5.04} & \B{11.42} & \B{24.13} \\
MISF~\cite{li2022misf} & GAN  & 14.52 & 3.58 & 16.23 & 18.05 & 2.8 & 13.19 & 9.92 & 4.27 & 14.53 & 21.04 & 0.43 & 1.88 \\
CoordFill~\cite{liu2023coordfill} & GAN  & 6.32 & 10.4 & 27.74 & 14.52 & 3.58 & 16.23 & 4.27 & 7.35 & 20.13 & 10.54 & 1.57 & 5.52 \\
ZITS~\cite{dong2022incremental} & GAN  & 4.25 & 19.56 & 34.56 & 8.24 & 10.84 & 25.74 & - & - & - & - & - & - \\
MAE-FAR~\cite{cao2022learning}                   & GAN & \B{4.06} & 22.18 & 36.82 & 7.71 & 15.06 & 28.74 & -    &  -    &  -    &  -    &  -   &  - \\
LaMa~\cite{lama}                 & GAN & 4.09 & 22.18 & 36.58 & 8.00 & 13.54 & 27.47 & 4.06 & 8.55  & 21.34 & 8.59 & 2.17  & 7.41 \\
CoModGAN~\cite{zhao2021comodgan} & GAN & 4.87 & 22.44 & 35.99 & 8.73 & 15.60 & 30.10 & - & - & - & - & - & - \\
PUT~\cite{put}                   & AR  & 7.73 & 2.68 & 18.35 & 15.17 & 2.54 & 12.89 & -    &  -    &  -    &  -    &  -   &  - \\
\bottomrule

				\end{tabular}
    }
    \vspace{-0.3cm}
    \caption{
      Quantitative comparisons on Places~\cite{places} and CelebA-HQ~\cite{pggan}.
      The \A{best} and \B{second best} results are in red and blue.
      Stable Diffusion inpainting model is trained on LAION-Aesthetics V2 5$+$~\cite{sd21inpaint}.
    }
    \label{tab:maintable}
  \end{table*}
}

\newcommand{\figablationcomparison}{
  \begin{figure*}
    \centering
    \includegraphics[width=\linewidth]{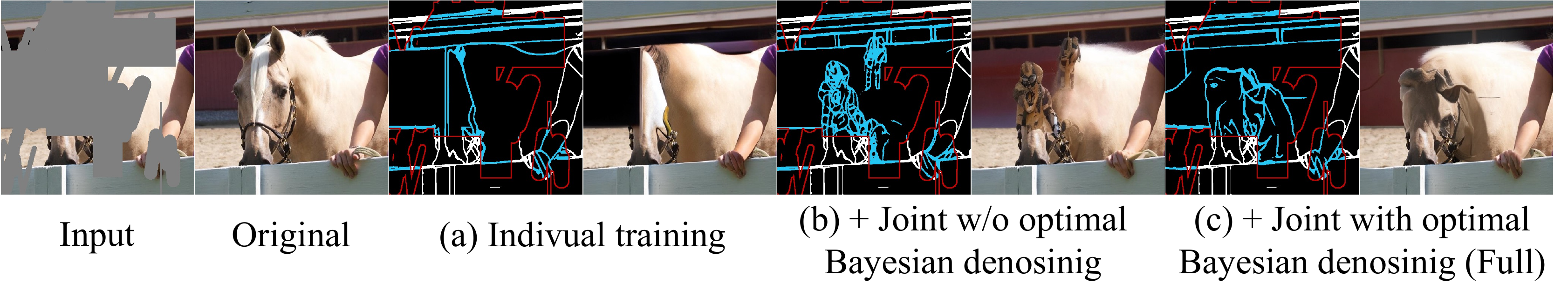}
    \vspace{-0.75cm}
    \caption{
      Visual comparison among the models of the ablation study using the same seed.
    }
    \label{fig:ablationvisual}
  \end{figure*}
}

\newcommand{\figmainfigure}{
  \begin{figure*}
    \centering
    \includegraphics[width=\linewidth]{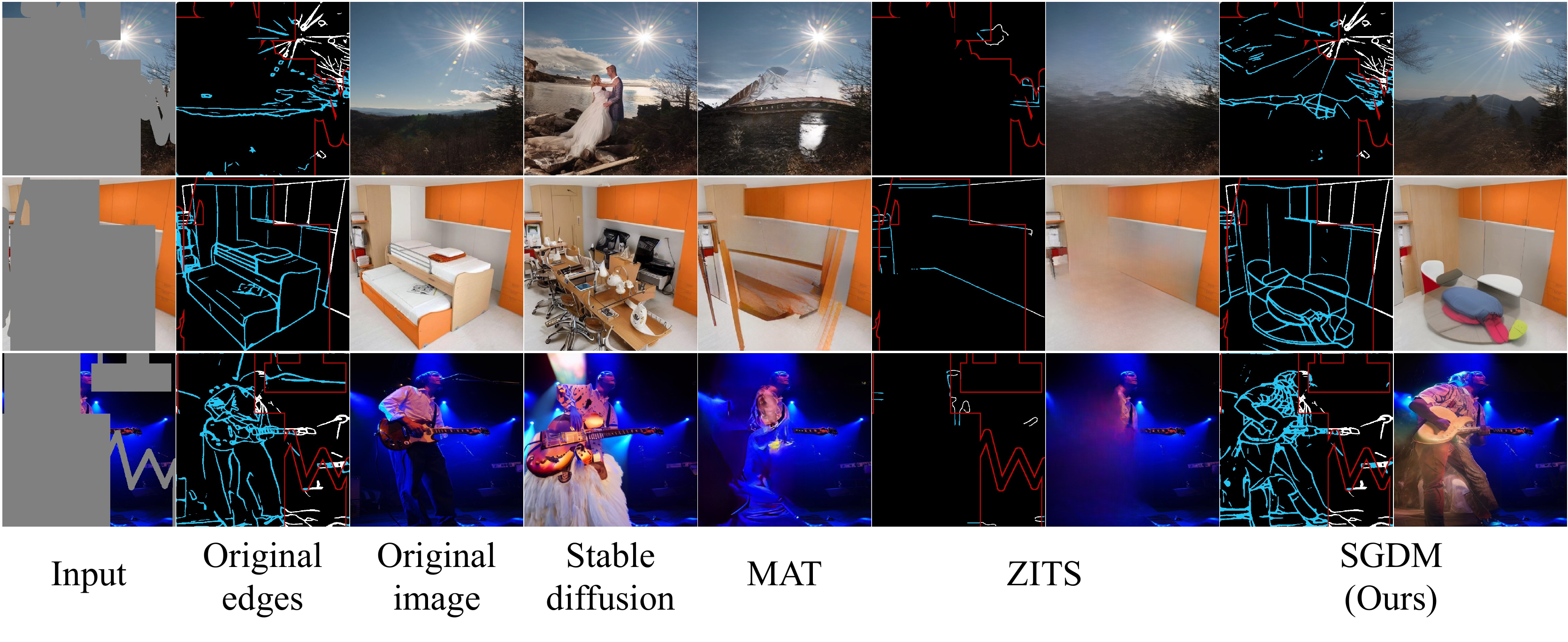}
    \vspace{-0.7cm}
    \caption{
      Qualitative comparison between existing methods and the SGDM on Places.
    }
    \label{fig:mainfigure}
  \end{figure*}
}

\newcommand{\figmaskratio}{
  \begin{figure}[h]
    \centering
    \includegraphics[width=\linewidth]{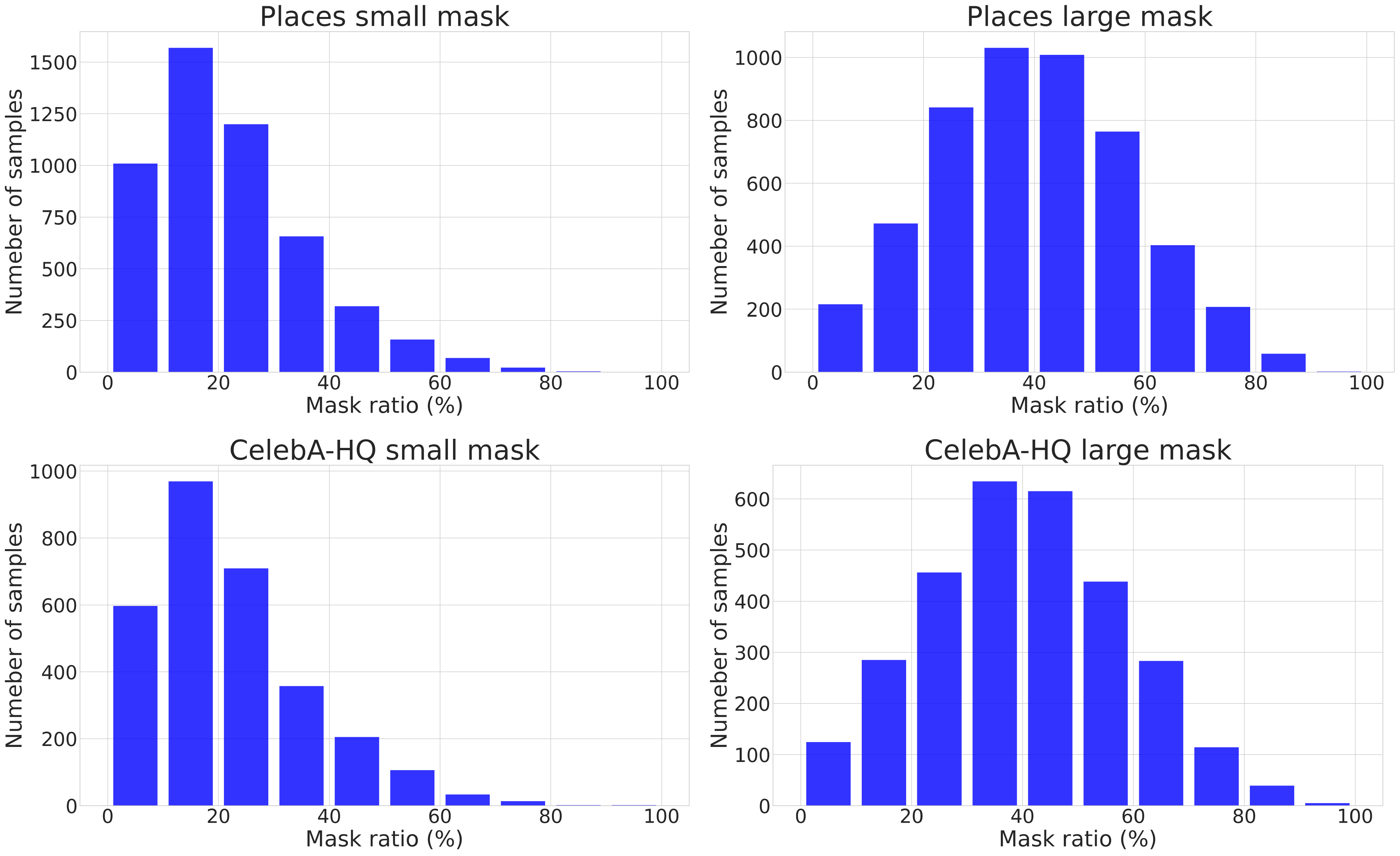}
    \caption{
       Statistics of large and small masks of Places and CelebA-HQ dataset for the quantitative evaluation.
    }
    \label{fig:figmaskratio}
  \end{figure}
}

\newcommand{\figdiverserock}{
  \begin{figure}
    \centering
    \includegraphics[width=\linewidth]{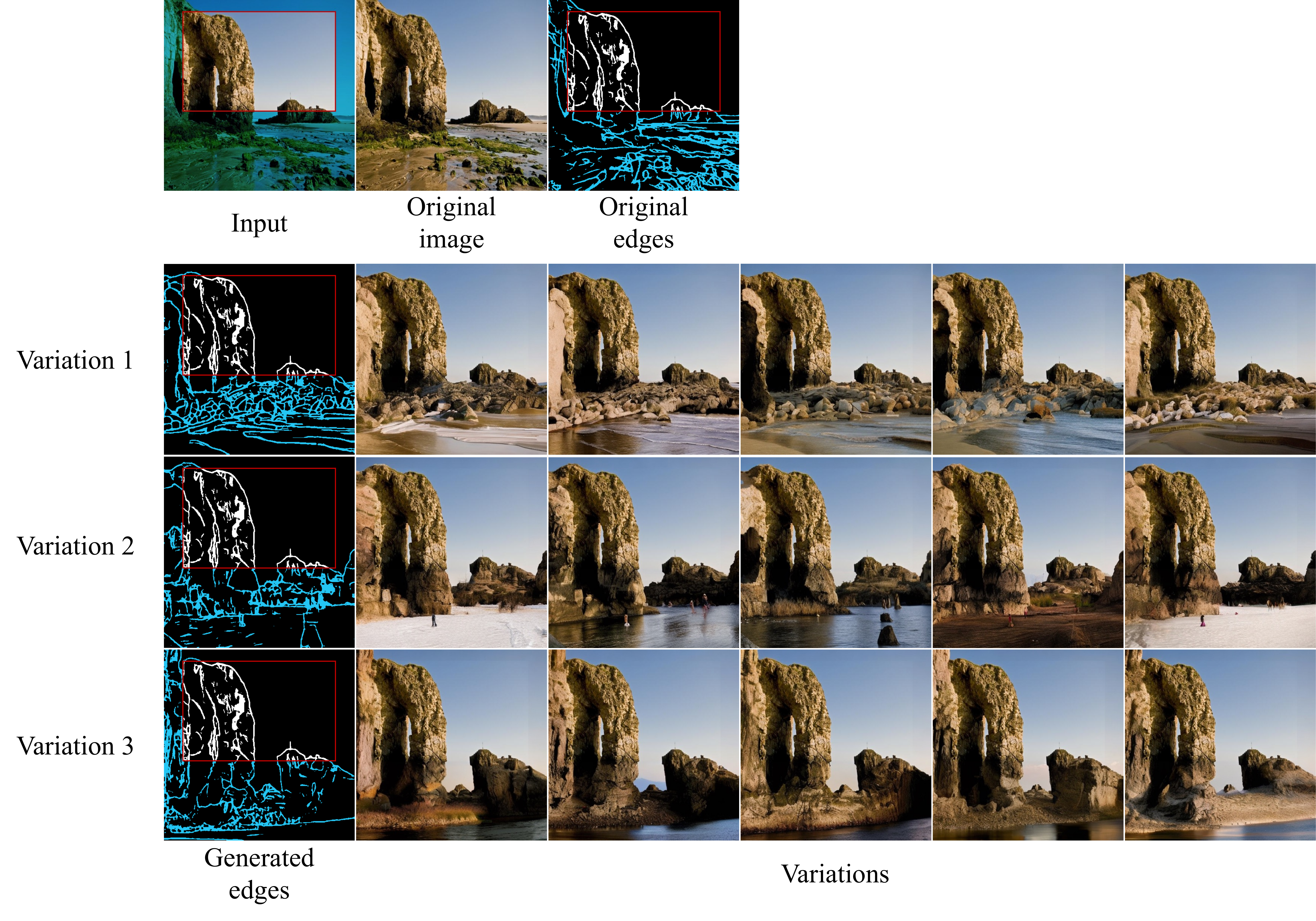}
    \vspace{-0.7cm}
    \caption{
        Qualitative comparisons of the generated different outputs by the SGDM.
    }
    \label{fig:figdiverserock}
  \end{figure}
}

\newcommand{\figsuppscelebahq}{
  \begin{figure}
    \centering
    \includegraphics[width=\linewidth]{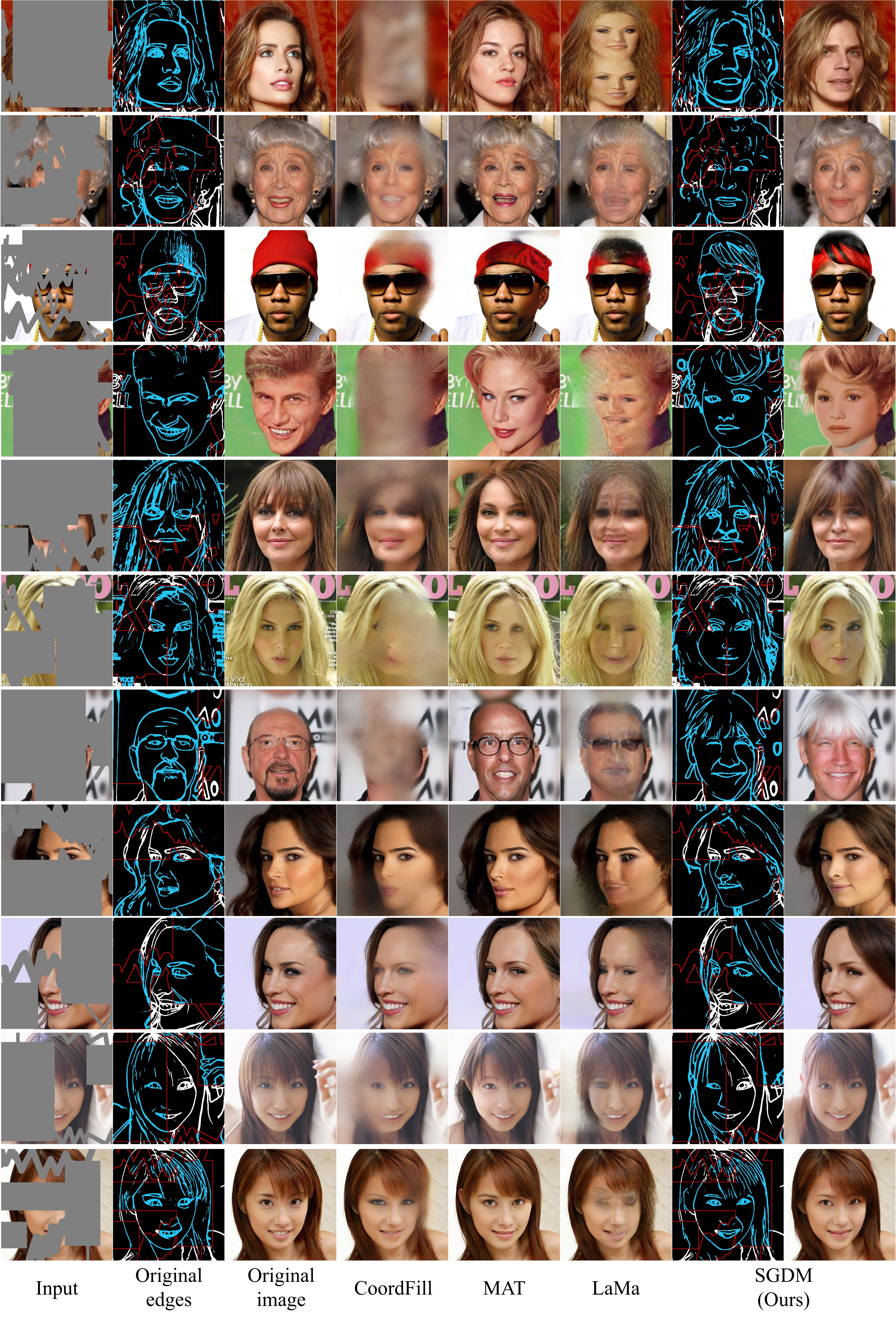}
    \vspace{-0.7cm}
    \caption{
        Qualitative comparisons of the proposed SGDM with the comparison methods using CelebA-HQ.
    }
    \label{fig:suppscelebahq}
  \end{figure}
}

\newcommand{\figsuppsplacesA}{
  \begin{figure}
    \centering
    \includegraphics[width=\linewidth]{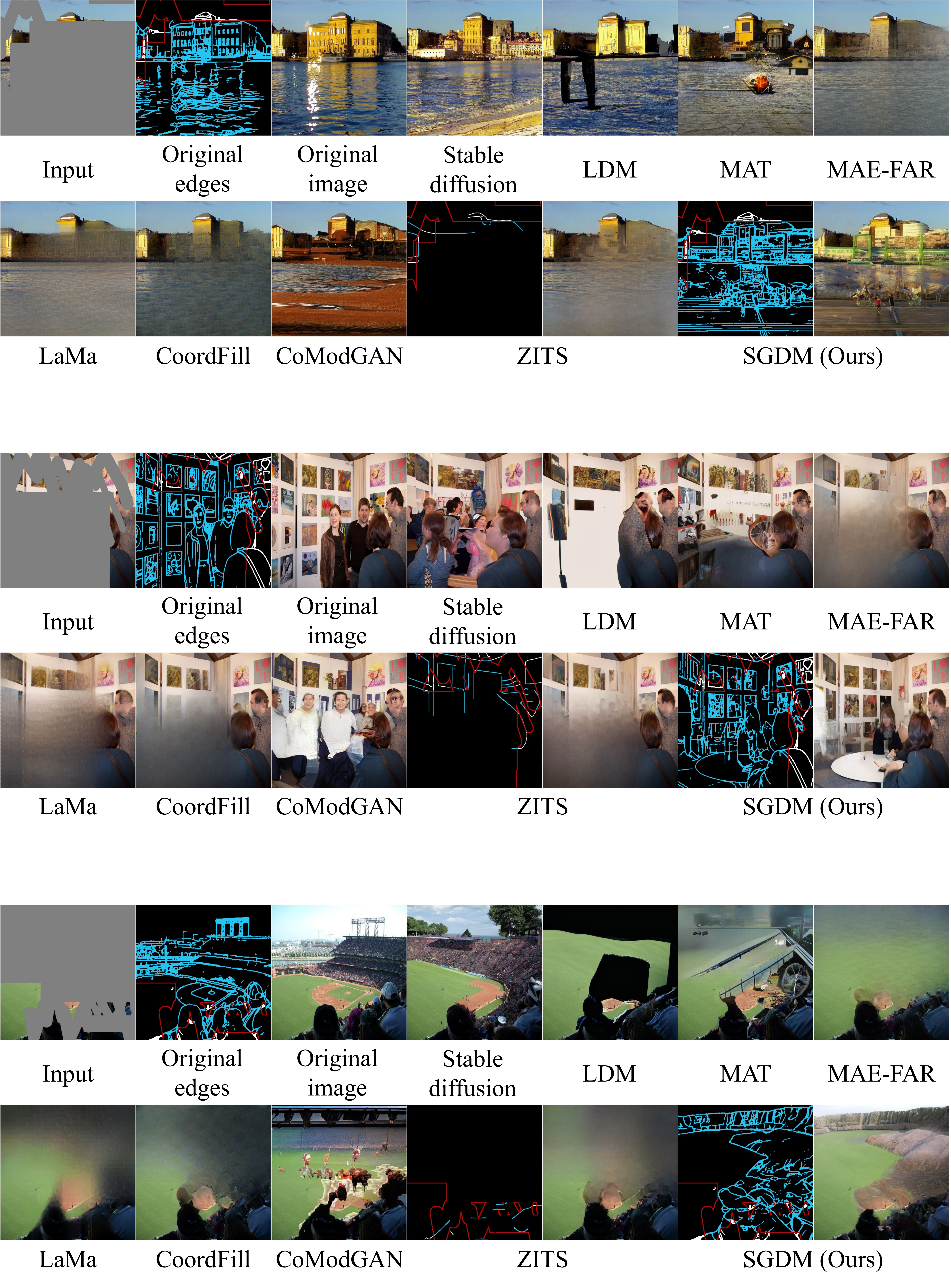}
    \vspace{-0.7cm}
    \caption{
        Qualitative comparisons of the proposed SGDM with the comparison methods using Places.
    }
    \label{fig:figsuppsplacesa}
  \end{figure}
}

\newcommand{\figsuppsplacesB}{
  \begin{figure}
    \centering
    \includegraphics[width=\linewidth]{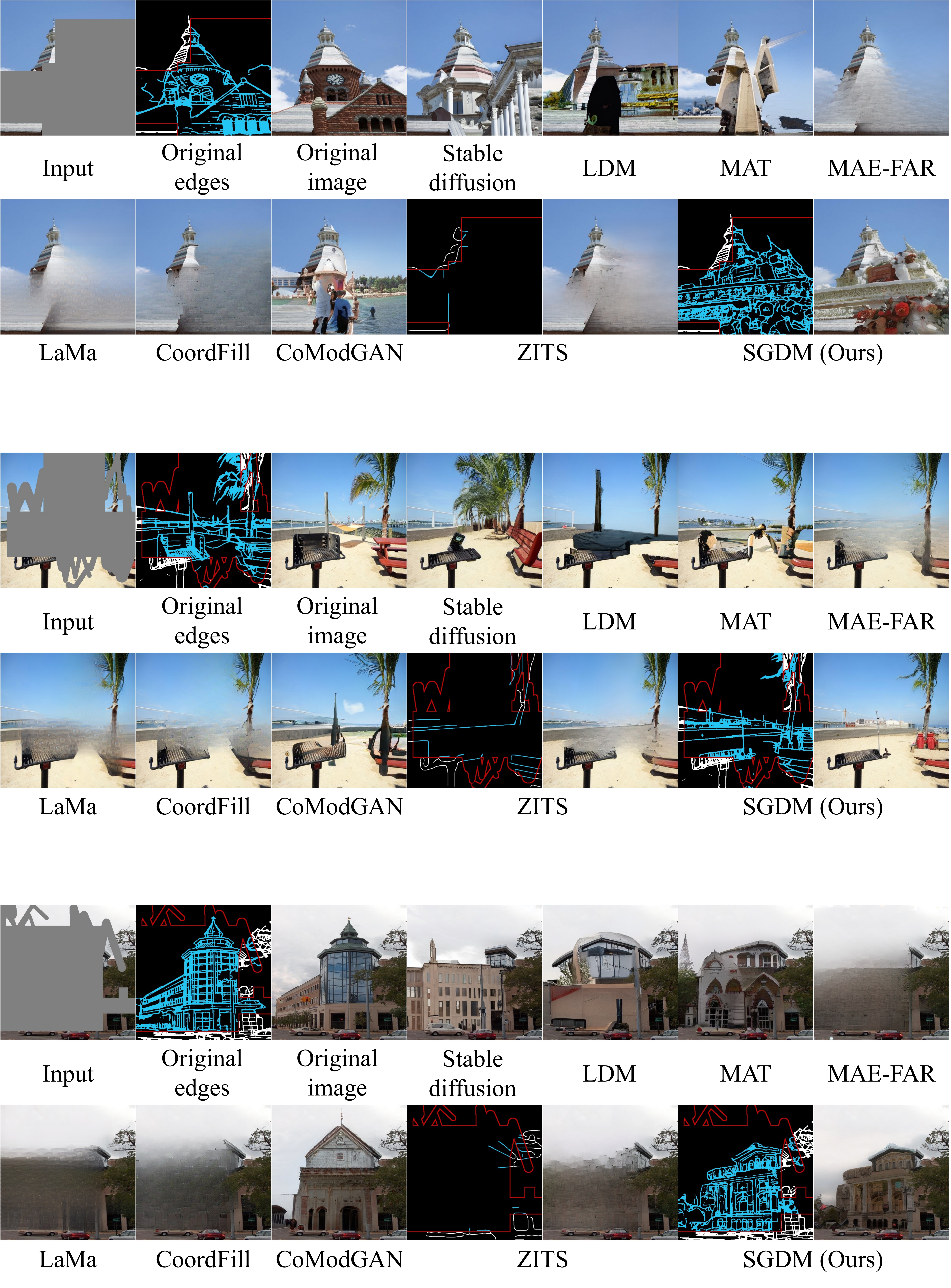}
    \vspace{-0.7cm}
    \caption{
        Qualitative comparisons of the proposed SGDM with the comparison methods using Places.
    }
    \label{fig:figsuppsplacesb}
  \end{figure}
}

\newcommand{\figpromptediting}{
  \begin{figure}
    \centering
    \includegraphics[width=\linewidth]{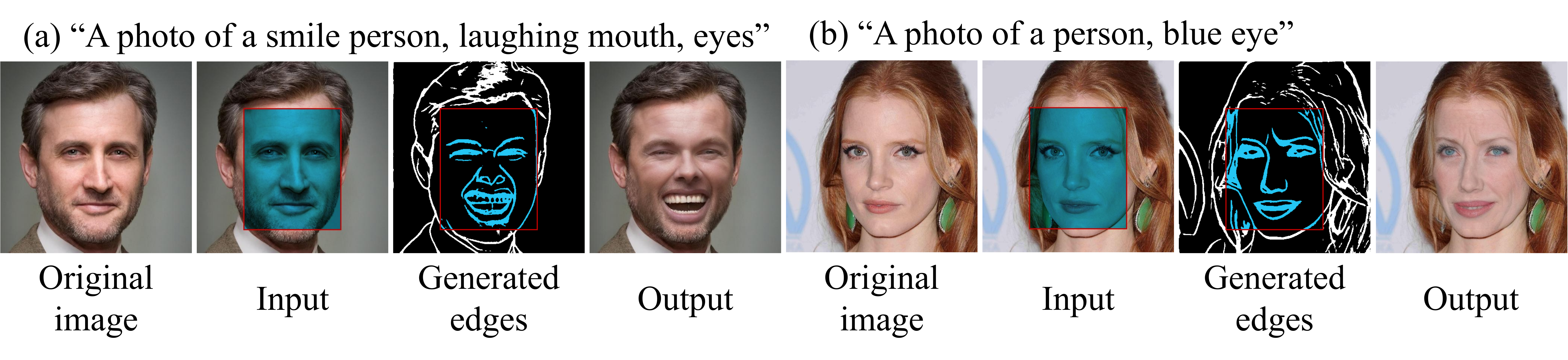}
    \vspace{-0.8cm}
    \caption{
        Language-guided image completion for (a) structure and (b) texture modifications.
    }
    \label{fig:promptediting}
  \end{figure}
}

\newcommand{\figrawedges}{
  \begin{figure}
    \centering
    \includegraphics[width=0.9\linewidth]{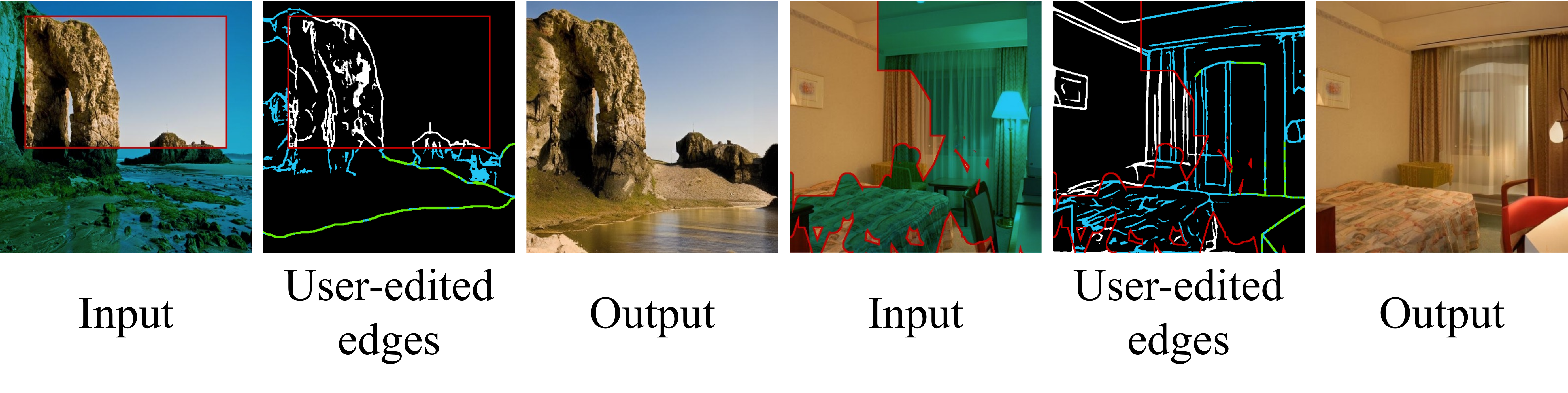}
    \vspace{-0.3cm}
    \caption{
        Image completion results using user-edited edges directly (without refinement).
        The input images and user-edited edge images are the same as in \cref{fig:teaser}.
    }
    \label{fig:rawedges}
    \vspace{-3mm}
  \end{figure}
}

%% file: abstract.tex

Image completion techniques have made significant progress in filling missing regions (\ie, holes) in images.
However, large-hole completion remains challenging due to limited structural information.
In this paper, we address this problem by integrating explicit structural guidance into diffusion-based image completion, forming our structure-guided diffusion model (SGDM).
It consists of two cascaded diffusion probabilistic models: structure and texture generators.
The structure generator generates an edge image representing plausible structures within the holes, which is then used for guiding the texture generation process.
To train both generators jointly, we devise a novel strategy that leverages optimal Bayesian denoising, which denoises the output of the structure generator in a single step and thus allows backpropagation.
Our diffusion-based approach enables a diversity of plausible completions, while the editable edges allow for editing parts of an image.
Our experiments on natural scene (Places) and face (CelebA-HQ) datasets demonstrate that our method achieves a superior or comparable visual quality compared to state-of-the-art approaches.
The code is available for research purposes at \url{https://github.com/UdonDa/Structure_Guided_Diffusion_Model}.

%% file: body.tex

\section{Introduction}
\label{sec:intro}

Image completion aims to fill missing regions (\ie holes) in images with visually coherent content.
Prior work has used guidance clues such as edges~\cite{Nazeri_2019_ICCV, Yu_2019_ICCV} and semantic maps~\cite{Liao_2021_CVPR, sgenet} to divide the problem into structure and texture generation.
These attempts have enabled various image-editing applications like object removal~\cite{sesami}, insertion~\cite{replace_inpaint}, and manipulation~\cite{Jo19iccv, zeng2022sketchedit}.
However, \emph{large-hole} completion remains a challenge because of the difficulty in generating useful guidance clues.

Image completion has been achieved using various techniques, including convolutional neural networks~\cite{NIPS2012_c399862d} and generative adversarial networks (GANs)~\cite{GAN, stylegan}.
While GAN-based methods~\cite{zhao2021comodgan, li2022mat} are good at filling large holes in flat texture patterns, they often produce images with distorted structures.
Researchers have tried to integrate structural guidances~\cite{Nazeri_2019_ICCV, Yu_2019_ICCV, dong2022incremental} to fill holes with rational structure;
however, they still struggle to generate reasonable clues for large holes.
Recently, autoregressive (AR) transformers~\cite{pixel_reccurent_nn, ict} and diffusion models (DMs)~\cite{ddpm, score_sde} have gained attention as promising techniques in image completion.
Their ability to generate diverse results is an additional strength from an application viewpoint.
Still, these techniques struggle to fill holes with coherent structures.
Thus, providing diverse ways to complete large missing regions with coherent structures remains challenging.

To address this challenge, we focus on DMs and explore the incorporation of structural guidance into the image completion process.
We propose a \emph{structure-guided diffusion model} (SGDM), which explicitly considers structural guidance using \emph{edge} information.
Our framework consists of a structure generator that generates plausible edges and a texture generator that completes textures guided by the edges.
Leveraging structural guidance and DMs, our SGDM can complete large holes with diverse and structurally coherent results.
Additionally, structural guidance provides opportunities for user-guided image editing, such as using sketching tools (see \cref{fig:teaser}).

To train the structure and texture generators simultaneously, we present a novel joint-training strategy using \emph{optimal Bayesian denoising}, specifically Tweedie's formula~\cite{miyasawa61, efron11, stein81}, which can denoise noisy images in a single step and thus allows backpropagation.
End-to-end training of two cascaded DM-based generators is a non-trivial problem; the SGDM's training is conditioned by noise, preventing the texture generator from directly using the noisy edge map from the structure generator.
We solve this problem by using optimal Bayesian denoising, improving the generalizability as in multi-task learning~\cite{mtl_overview, mtl_stitch}.
Our experiments with natural scene (Places~\cite{places}) and face (CelebA-HQ~\cite{pggan}) datasets demonstrate that our method achieves a superior or comparable visual quality compared to state-of-the-art methods.

Our contributions are summarized as follows.
1) We propose the structure-guided diffusion model (SGDM) for large-hole image completion.
It consists of structure and texture generators producing coherent, realistic contexts in large holes.
As far as we know, this is the first work combining structural generation and guidance with diffusion models for image completion.
2) We design a novel joint-training strategy using optimal Bayesian denoising to enable end-to-end training of two cascaded DM-based generators.
3) We show that SGDM achieves state-of-the-art or comparable visual quality on both Places~\cite{places} and CelebA-HQ~\cite{pggan} datasets.

\section{Related Work}
\label{sec:related_works}

\subsection{Deterministic Image Completion}

The advent of deep learning brought significant success to image completion, especially GAN-based methods~\cite{iizuka17, zheng2022cmgan}.
To achieve fine-grained textures, many works proposed task-specific operations such as global and local discriminators~\cite{iizuka17}, attention mechanisms~\cite{Yu18, yan2019PENnet, Liu_2019_CSA, xie19iccv}, partial~\cite{partial_conv}, and gated~\cite{Yu_2019_ICCV} convolutions.
Concurrently, several works utilized explicit clues such as object edges~\cite{Nazeri_2019_ICCV, dong2022incremental, cao21sketch}, foreground contours~\cite{foreground_inpainting, zeng2022sketchedit, Jo19iccv}, smoothed images~\cite{ren2019structureflow}, reference images~\cite{Zhou_2021_CVPR}, confidence maps~\cite{zeng2020highresolution}, and semantic segmentation maps~\cite{Liao_2021_CVPR, sgenet}.
Nazeri \etal~\cite{Nazeri_2019_ICCV} first proposed a two-stage framework for edges and textures, introducing structure guidance.
ZITS~\cite{dong2022incremental} used an attention-based transformer~\cite{vaswani17} to predict structural guidance.
However, these methods still have difficulties in predicting guidance within large missing regions.
In contrast, the use of DMs with strong model capacity and novel joint training generates reasonable edges.

\subsection{Diverse Image Completion}
Recent image completion studies have addressed more challenging issues, \ie, filling large holes in images with multiple visually plausible and diverse contents~\cite{Peng_2021_CVPR, Liu_2021_CVPR, pluralistic_free_from, li2023image, normflow_inpaint}.
Variational-auto-encoder-based methods~\cite{zheng2019pluralistic, uctgan} demonstrated diverse image completion, although their synthesized quality was limited due to variational training~\cite{vae_understandings}.
Subsequently, CoModGAN~\cite{zhao2021comodgan} and MAT~\cite{li2022mat} successfully filled large holes, particularly in flat texture patterns.
However, they often produce images with distorted or unrealistic structures and limited diversity.

Recent studies~\cite{batfill,ict,put} have focused on an AR transformer~\cite{vaswani17}, achieving high-fidelity quality and diversity.
However, AR-based methods~\cite{ict, batfill} faced information loss issues due to ``low resolution'' and ``quantization,'' which down-sample images to a much lower resolution (\eg $32\times32$) and quantize RGB values $256^3$ into a much lower dimension 512.
To address this, PUT~\cite{put} used a patch-based autoencoder with VQVAE~\cite{vqvae} and applied the AR transformer to vector quantized tokens.
Nonetheless, AR transformer-based approaches struggle with handling high-resolution images and sampling orders~\cite{esser2021imagebart} due to their pixel-by-pixel autoregressive sampling.
In contrast, our proposed method overcomes these limitations by leveraging diffusion models and explicit structural guidance, which allows for generating coherent structures and diverse images.

\subsection{Image Completion with Diffusion Models}
Diffusion and score-based models have emerged as a family of likelihood-based models, showing remarkable success in quality, diversity, mode coverage, and generality in their training objective~\cite{song19, ddpm}.
Most previous studies~\cite{score_sde, jascha15, repaint} have demonstrated image completion using unconditional image generation models by replacing the known region with a designated hole at each sampling step.
However, a major limitation of these methods is their inability to produce harmonious images that match the known regions.
To address this, Palette~\cite{palette} learned conditional completion, and RePaint~\cite{repaint} introduced a conditional sampling method, which alternately performs the forward and reverse diffusion processes for pre-trained unconditional models.
Nonetheless, these methods often generate irrelevant content for large holes.
Our method overcomes this limitation by explicitly estimating the structure of missing regions and using it as guidance.
ControlNet~\cite{zhang2023adding} proposed an encoder to support additional input conditions for pretrained diffusion models.
It also uses an edge map as structural guidance to fill holes.
However, edges within holes must be prepared beforehand.
In contrast, our method learns to generate edges within missing regions and then synthesize textures.

\section{Preliminaries}
This section provides an overview of diffusion models, focusing on denoising diffusion probabilistic models (DDPM) and latent diffusion models (LDM), which underpin our proposed method.
We also introduce optimal Bayesian denoising, which is crucial for our joint training.

\subsection{Diffusion Models}
\label{subsec:diffusion_models}

DDPM is built upon a discrete Markov chain between two processes: forward and reverse processes.
The forward process, initiated from a noiseless data $x_{0}$, adds Gaussian noise to a previous data $x_{t-1}$ at each timestep $t$ to generate a current data $x_{t}$.
The reverse process, on the other hand, iteratively samples $x_{t-1}$ from $x_{t}$, starting from a Gaussian noise $x_{T}$.
The forward process is modeled as $q(x_{t} | x_{t-1}) \coloneqq \mathcal{N}(x_{t} ; \sqrt{1 - \beta_{t}} x_{t-1}, \beta_{t} \mathbf{I} )$, where $\beta_{t}$ is a pre-defined noise scale depending on timestep $t$.
By accumulating the timesteps, we can express the forward to any time $t$ from a data as a reparameterization trick~\cite{kingma2022autoencoding}:
$q(x_{t} | x_{0}) = \mathcal{N}( x_{t} ; \sqrt{\alpha_{t}} x_{0}, (1 - \alpha_{t}) \mathbf{I} )$,
where $\alpha_{t} \coloneqq \prod\nolimits_{i=1}^t (1 - \beta_{i}) $.
The reverse process is modeled as $ p_{\theta} (x_{t-1} | x_{t}) \coloneqq  \mathcal{N} (x_{t-1} ;  \mu_{\theta} ( x_{t}, t), \Sigma_{\theta} (x_{t}, t) )$.
DDPM employs UNet~\cite{unet} to approximate this posterior.
For conditional cases, the UNet $\epsilon_{\theta}$ can be trained using a denoising score matching $\Vert \epsilon_{\theta}(x_{t}, t, y) - \epsilon \Vert^{2}_{2}$, where $\epsilon$ is a Gaussian noise added to $x_{0}$ to create $x_{t}$, $y$ is a condition such as an image and an edge image with a hole in our case, and $\epsilon_{\theta}(x_{t}, t, y)$ represents a score function of the perturbed data distribution, $\nabla_{x_{t}} \log p_{\theta} (x_{t})$.
After the training, DDPM can generate images using annealed Langevin dynamics~\cite{song19}.

\subsection{Latent Diffusion Model}
\label{subsec:ldm}
DMs typically learn denoising in an RGB pixel space, which requires high-computational costs for high-resolution images.
In contrast, LDM~\cite{ldm} learns denoising in the latent space of pretrained autoencoders, significantly improving both training and sampling efficiency without compromising quality compared to pixel-based DMs.

LDM consists of an autoencoder with an encoder $\mathcal{E}$ and decoder $\mathcal{D}$ in an RGB space as well as denoising autoencoders in the latent space.
Given a data $x \in \mathbb{R}^{H \times W \times 3}$ in an RGB space, the encoder $\mathcal{E}$ encodes $x$ into a latent representation $z = \mathcal{E}(x)$, and the decoder $\mathcal{D}$ reconstructs the data from the latent $z \in \mathbb{R}^{h \times w \times c}$, denoted as $\tilde{x} = \mathcal{D}(z) = \mathcal{D}(\mathcal{E}(x))$, where $h$ and $w$ are downsampled latent size.
LDM learns the denoising autoencoder $\epsilon_{\theta}$ in the latent space using DDPM.

\subsection{Optimal Bayesian Denoising}
Optimal Bayesian denoising is a technique that performs minimum mean square error (MMSE) denoising in a single step.
Given a Gaussian noise $\epsilon \sim \mathcal{N} (\epsilon; \boldsymbol{\mu}, \boldsymbol{\Sigma}) $, its MMSE estimator is given by Tweedie's formula~\cite{stein81, efron11, miyasawa61}; that is, $\mathbb{E}[ \boldsymbol{\mu} | \epsilon ] = \epsilon + \boldsymbol{\Sigma} \nabla_{\epsilon} \log p(\epsilon) $.
In DDPM, the forward step is modeled as $q(x_{t} | x_{0}) = \mathcal{N}( x_{t} ; \sqrt{\alpha_{t}} x_{0}, (1 - \alpha_{t}) \mathbf{I} )$ as described in \cref{subsec:diffusion_models}.
Therefore, we can apply Tweedie's formula by substituting $\sqrt{\alpha_{t}} x_{0}$ and $(1 - \alpha_{t}) \mathbf{I}$ for $\boldsymbol{\mu}$ and $\boldsymbol{\Sigma}$, respectively.
This allows us to determine a single-step denoising operation as
\begin{align}
  \label{eq:tweedie}
  F(x_{t}) \coloneqq \hat{x}^{t}_{0} = \frac{ x_{t} +  (1 - \alpha_{t}) \nabla_{x_{t}} \log p(x_{t})  }{ \sqrt{\alpha_{t}} },
\end{align}
where $\hat{x}^{t}_{0}$ represents a denoised sample.
This operation enables us to convert the noisy sample (at time $t$) into a denoised one (at time $0$) in a single step, provided that the optimal score function $\nabla_{x_{t}} \log p(x_{t})$ is known.
To reconstruct a denoised sample in an RGB space, the operation is written as $\mathcal{D}(F(x_{t}))$; however, we will omit the decoder $\mathcal{D}$ for simplicity.
We note that previous studies have used the single-step denoising technique for sampling~\cite{song2020ddim} or a formulation of DMs~\cite{karras2022elucidating} while we aim to use it for a component of training.

\section{Structure-Guided Diffusion Model}
\label{sec:approach}

Given an input image with missing regions (\ie holes), our goal is to generate a structurally reasonable image that respects the context of the visible regions.
We denote the target image by $I \in \mathbb{R}^{H \times W \times 3}$, the binary mask representing the missing regions by $M \in \{0, 1\}^{H \times W \times 1}$, and the generated image by $\hat{I} \in \mathbb{R}^{H \times W \times 3}$, where $H$ and $W$ represent a spatial resolution.
With this notation, the objective is to generate $\hat{I}$ from $I_{M} = I \odot M$.
Our proposed SGDM utilizes structural guidance during the generation process.
Specifically, it generates a hole-filled edge image $\hat{E}$ and then uses it as structural guidance to generate the output image $\hat{I}$.
The edge image $\hat{E}$ is generated using an edge image with missing regions, denoted by $E_{M}$, which is produced from $I_{M}$ using an existing edge detection algorithm~\cite{soria2023dexined_ext}.

\figindivjointtile

\subsection{Framework Architecture}

Our framework consists of two DM-based networks: a structure generator $f_{\theta}$ and a texture generator $g_{\phi}$, where $\theta$ and $\phi$ are learnable parameters.
In particular, we employ LDM~\cite{ldm} to reduce computational cost in high-resolution images as described in \cref{subsec:ldm}.
We train a condition encoder $R$ to encode input conditions with five channels for each network, including a mask, masked image, masked edge image, and hole-filled edge image, using~\cite{zhang2023adding}.
The encoded condition is incorporated into the generator.

The framework procedure for hole-filling follows these steps:
First, the structure generator fills in the holes of the edge image $E_{M}$ to produce the hole-filled edge image $\hat{E}$.
Then, the texture generator creates plausible textures with the guidance of $\hat{E}$ while maintaining the context of the visible regions of the input image.
These generations use the iterative sampling of DMs as described in \cref{subsec:diffusion_models}.
Compared to GAN and AR methods, the SGDM using DMs has two advantages: 1) input data with noise have no holes and 2) iterative generation.
In other words, the SGDM generates an output by recursively denoising a noisy input without holes, whereas GAN and AR methods predict the output from a masked input that lacks most contextual information.

\subsection{Individual Training}
\label{subsec:1st_training}

We describe the data preparation and training procedure.
Suppose we have a ground-truth image $I$ from a training dataset.
Then, we extract an edge image $E$ using the edge detection algorithm~\cite{soria2023dexined_ext}.
We degrade the image $I$ and the edge image $E$ using a binary mask $M$, which is randomly drawn for each sample, denoted as the masked image $I_{M} = I \odot M$ and the masked edge image $E_{M} = E \odot M$, respectively.
We fill the masked region with zeros.
To train DMs, we create a noisy image $I_{t} \in \mathbb{R}^{h \times w \times 3}$ and a noisy edge image $E_{t} \in \mathbb{R}^{h \times w \times 3}$ in a latent space at timestep $t$ (out of $T$ timesteps) using Gaussian noises $\epsilon_{I}$ and encoded image $\epsilon_{E}$ with edge map $\mathcal{E}(I)$ and $\mathcal{E}(E)$, respectively, following the reparameterization trick in \cref{subsec:diffusion_models}.

\Cref{fig:framework} illustrates our individual training process.
The structure generator $f_{\theta}$ generates, from the noisy edge image at timestep $t$, a less noisy edge image at the previous timestep $t-1$.
More specifically, given the noisy edge image $E_{t}$, masked image $I_{M}$, mask $M$, masked edge image $E_{M}$, and timestep $t$, it outputs a less noisy edge image $\hat{E}_{t-1} \in \mathbb{R}^{h \times w \times 3}$.
Similarly, given the noisy image $I_{t}$, masked image $I_{M}$, mask $M$, edge image (without noise or masked regions) $E$, and timestep $t$, the texture generator $g_{\phi}$ outputs $\hat{I}_{t-1} \in \mathbb{R}^{h \times w \times 3}$.
These processes can be written as
\begin{align}
  f_{\theta} ( E_{t}, R_{\hat{\theta}}(I_{M}, M, E_{M}), t) = \hat{E}_{t-1},~
  g_{\phi} ( I_{t}, R_{\hat{\phi}}(I_{M}, M, E), t)         = \hat{I}_{t-1},
  \label{eq:texture_generator}
\end{align}
where $\hat{\theta}$ and $\hat{\phi}$ are parameters of condition encoders $R$ for the structure $f_{\theta}$ and texture generator $g_{\phi}$, respectively.
These networks can be trained via the denoising score matching~\cite{ddpm} in a closed form as described in \cref{subsec:diffusion_models},
\begin{align}
  \mathcal{L}_{f} & =  \mathbb{E}_{I, M, E, t, \epsilon_{E}} \big[ \Vert f_{\theta}(E_{t}, R_{\hat{\theta}}(I_{M}, M, E_{M}), t) - \epsilon_{E} \Vert^{2} \big],      \\
  \mathcal{L}_{g} & = \mathbb{E}_{I, M, E, t, \epsilon_{I}} \big[ \Vert g_{\phi}(I_{t}, R_{\hat{\phi}}(I_{M}, M, E), t) - \epsilon_{I} \Vert^{2} \big] \label{eq:mse},
\end{align}
where the noises $\epsilon_{E}$ and $\epsilon_{I}$ are sampled from Gaussian distribution to create $E_{t}$ and $I_{t}$.

\subsection{Joint Training}

The individually trained structure generator sometimes generates unreasonable edges.
This is because edge images are sparse compared to textured images, making the modeling more difficult.
To mitigate this issue and improve generalization, we propose joint fine-tuning of both networks in an end-to-end manner after the individual training.
For this purpose, we propose a novel joint-training strategy using optimal Bayesian denoising, as shown in \cref{fig:framework}.

The joint training of DMs cannot be performed in a straightforward manner, such as in the training of GANs~\cite{Nazeri_2019_ICCV, dong2022incremental}.
This is because the texture generator requires a \emph{noiseless} edge image as input, but the structure generator cannot generate a noiseless edge image without iterative sampling.
Even if we produce a noiseless edge image via iterative sampling, backpropagation becomes intractable due to gradient accumulation and computational costs.
We tackle this issue by applying the single-step denoising operation in \cref{eq:tweedie};
that is, we obtain a noiseless estimate by $\hat{E}^{t}_{0} = F(E_{t})$.
This approach allows us to perform backpropagation in an end-to-end manner.
Finally, we formulate our total loss for the joint training as
\begin{align}
  \mathcal{L}_{jt} = \mathcal{L}_{f} + \mathcal{L}_{g\_o} + \mathcal{L}_{g\_d},
\end{align}
where $\mathcal{L}_{g\_o}$ and $\mathcal{L}_{g\_d}$ are both calculated by \cref{eq:mse}, but with different edge images.
$\mathcal{L}_{g\_d}$ uses edge images generated by the structure generator (\cref{eq:texture_generator}) and Tweedie's formula (\cref{eq:tweedie}), encouraging the structure generator to learn texture-aware edge prediction (see \cref{fig:ablationvisual}).
$\mathcal{L}_{g\_o}$ uses original (ground truth) edge images, regularizing the texture generator and preventing overfitting to edge images generated by the structure generator and Tweedie's formula.

\tabfaces

\section{Experiments}
\label{sec:experiment}

\paragraph{Datasets.}
The experiments were conducted with Places~\cite{places} and CelebA-HQ~\cite{pggan}, which cover different degrees of context (natural scenes only vs\onedot face).
The image resolution was $512 \times 512$ for all experiments.
For Places, we prepared a train set and a test set with 8 million (M) and 5,000 images.
The test set was created from the official test set for our evaluation.
For CelebA-HQ, we prepared a train set and a test set with 24,183 and 2,993 images, respectively.
For a better understanding of the performances for holes with various sizes, we prepared two different masks (\ie large and small masks) following MAT~\cite{li2022mat}.

\paragraph{Evaluation metrics.}
Following~\cite{li2022mat, zhao2021comodgan}, we used FID~\cite{fid}, P-IDS, and U-IDS~\cite{zhao2021comodgan} to measure a perceptual fidelity between ground truth and hole-filled images for evaluation.
P-IDS and U-IDS robustly assess perceptual fidelity and correlate well with human preferences~\cite{zhao2021comodgan}.
Similarity-based metrics such as PSNR and SSIM fail to measure completion, thus, we did not use these metrics.

\paragraph{Implementation details.}
Before the training, we initialized our generators' weights with stable-diffusion-2-1-base~\cite{sd21base}, which was trained using LAION-5B dataset~\cite{schuhmann2022laionb}.
We did not use any prompt inputs.
For the individual training, each network was trained for 25M images on Places and CelebA-HQ.
Additionally, we carried out the joint training with 1M images.
The batch size was fixed to 1.
Both trainings were performed with AdamW optimizer~\cite{adamw} with $\beta_{1} = 0.9$ and $\beta_{2} = 0.999$ and a learning rate of $10^{-5}$.
We conducted all experiments with four NVIDIA A100 GPUs.
To generate images, we used RePaint~\cite{repaint} sampler.

\subsection{Comparison with State-of-the-Art Methods}
\paragraph{Quantitative comparisons.}
We provide the quantitative performance with different masked regions on Places and CelebA-HQ, respectively.
Only ZITS~\cite{dong2022incremental} used an edge map as structure guidance.
\Cref{tab:maintable} shows the SGDM achieved the best performance in all metrics under both small and large masks on CelebA-HQ.
However, on Places, the SGDM yielded the best FID, but demonstrated P-IDS and U-IDS comparable to MAT.

\paragraph{Qualitative comparisons.}
\Cref{fig:mainfigure} shows the qualitative comparison of the competing methods.
We see that the proposed SGDM was able to produce rational edges and coherent textures in large holes.
Stable diffusion (SD) and MAT produced content without blurring but generated messy results, especially for the region of the human and tables.
We observed that SD often generated persons unrelated to the context of the input.
ZITS failed to generate reasonable edges for large holes and the generated textures contain a lot of blur.
All these results demonstrate that the SGDM was superior to the current state-of-the-art methods.

\tabablationva

\figmainfigure

\figablationcomparison

\figrawedges

\subsection{Ablation Study}
\label{sec:ablation_study}
We conducted an ablation study with large masks using FID and LPIPS~\cite{lpips} to evaluate the image quality.
We created a subset with 1,000 images from the test set on Places.
We compared SGDMs with different settings: (a) only individual training with 25M training images, (b) joint training after the individual training with 0.1M images without optimal Bayesian denoising, and (c) joint training after individual training with optimal Bayesian denoising using 0.1M and 1M images as our full model.
\Cref{tab:ablation_study} shows the result.
Comparing models (b) and (c), we see that the joint training with optimal Bayesian denoising improved the metrics.
\Cref{fig:ablationvisual} visually compares the settings.
Model (a) could not generate reasonable edges and textures.
Model (b) could generate edges of the animal's head but failed to synthesize visually coherent textures.
In contrast, model (c) could generate realistic content.
We conjecture that the texture generator in model (b) did not learn contextual correspondences between edges and textures well because it was conditioned by noisy edges in training.
Joint training with optimal Bayesian denoising effectively improved the image quality.

\subsection{Applications}
\paragraph{Sketch-guided image completion.}
\Cref{fig:teaser} shows image completion results using user-edited edge images, highlighting SGDM's potential of use as a user-guided image editing tool.
First, the edge images generated by our structure generator (the third column) were manually edited by the user with sketching tools (the fifth column).
Then, they were further refined using the prior of the structure generator by SDEdit~\cite{meng2022sdedit} (the sixth column);
specifically, we first perturbed the user-edited edge images with Gaussian noise at timestep 500 and 200 out of 1,000, respectively, and then progressively removed the noise via the reverse process.
This process, called SDEdit, can refine (potentially) unrealistic user-edited edges, making them more compatible with the texture generator.
We found this approach to be sufficiently robust and well-suited for interactive editing.
Note that our SGDM can also generate plausible images directly from raw edges manually drawn by users, \ie without the refinement by SDEdit, as demonstrated in \cref{fig:rawedges}.

\paragraph{Language-guided image completion.}
\Cref{fig:promptediting} demonstrates a language-guided image completion application where user-provided text prompts successfully control structure and texture generation.
Note, however, that SGDM may not work well with more complex prompts because SGDM has not been explicitly trained using prompt input.

\figpromptediting

\section{Limitation and Discussion}

\paragraph{Failure cases.}
Our method sometimes struggled to generate structurally rational edges.
For example, in the second row of \cref{fig:teaser}, the auto-generated edges were not very rational, especially in the bottom right region.
Nonetheless, users can manually correct such irrational edges if necessary;
this flexibility is one of the SGDM's strengths.

\paragraph{Computational costs.}
Our method requires the iterative denoising process.
In contrast, GANs can generate images in a single step, meaning that our method inherently takes more time than GAN-based methods.
ARs-based methods such as PUT also require iterative inference, but their computational cost is lower than ours.
To investigate the computational cost issue, we measured the time needed to complete a center-masked image for each method.
As a result, MAT (GAN-based) needed 0.098 seconds, and PUT (AR-based) required 4.06 seconds, respectively.
Our method with the RePaint sampler, on the other hand, took 133 seconds.

\paragraph{Potential societal impacts.}
Our method inherits the potential societal impact of previous image completion methods (\eg \cite{repaint}).
Generated images may reflect the biases in the datasets, such as gender, age, ethnicity, \etc.
Moreover, the image-editing capability of our method could aid in DeepFake creation~\cite{deepfake1, deepfake2}.
On the other hand, image completion may enhance privacy protection by removing identifiable information from public-space photographs.

\paragraph{Future work.}
Even though our SGDM was not explicitly trained using prompts, our SGDM could reasonably perform language-guided image completion for simple prompts, as shown in \cref{fig:promptediting}.
The original ControlNet~\cite{zhang2023adding}, on the other hand, presents a training method that explicitly uses prompt input to achieve more sophisticated language guidance.
Exploring such training methods that utilize prompt input is a promising direction for future research.

\section{Conclusion}
We have presented a structure-guided diffusion model (SGDM), which uses structural guidance in image completion.
We have proposed a novel training strategy to enable effective end-to-end training.
Extensive experiments show that the SGDM achieves a superior or comparable visual quality on both Places and CelebA-HQ as compared to state-of-the-art methods.
Incorporating structural guidance has not only improved the visual quality but also enabled user-guided image editing.

\subsection*{Acknowledgement} 
This research was partly supported by JSPS KAKENHI 22KJ1014, JSPS KAKENHI 21H03460, and D-CORE Grant from Microsoft Research Asia.
We would like to thank Seiji Kurokoshi and Atsushi Honda for the helpful discussions.

%% file: supps_body.tex

\clearpage
	
\renewcommand\thesection{\Alph{section}}
\renewcommand\thesubsection{\thesection.\arabic{subsection}}
\renewcommand\thefigure{\Alph{section}.\arabic{figure}}
\renewcommand\thetable{\Alph{section}.\arabic{table}} 

\setcounter{section}{0}
\setcounter{figure}{0}
\setcounter{table}{0}

\begin{center}
    \noindent{\Large{\textbf{Appendix}}}
\end{center}

\newcommand\beginsupplement{%
        \setcounter{table}{0}
        \renewcommand{\thetable}{\Alph{table}}%
        \setcounter{figure}{0}
        \renewcommand{\thefigure}{\Alph{figure}}%
        \renewcommand\thesection{\Alph{section}}
        \renewcommand\thesubsection{\thesection.\alph{subsection}}

     }
\beginsupplement

\section{Mask Statistics}

We followed MAT~\cite{li2022mat} and DeepFill v2~\cite{Yu_2019_ICCV} to create masks.
\Cref{fig:figmaskratio} show statistics of large and small masks of Places~\cite{places} and CelebA-HQ~\cite{pggan} dataset for the quantitative evaluation.
In the training, we created zero to five full-size or half-size rectangles and zero to nine random strokes with 12 to 48 width and 4 to 18 vertex.
In the evaluation, we additionally created small masks.
For the small masks, we created zero to three full-size or half-size rectangles and zero to four random strokes with 12 to 48 width and 4 to 18 vertex.
For the center masks, we removed the top and bottom center area of the 512 $\times$ 512 images from 136 to 394 pixels.

\section{Additional Qualitative Comparisons}

We present more qualitative comparisons of the proposed SGDM with comparison methods.
\Cref{fig:figdiverserock} show other variations of Fig. 1.
\Cref{fig:figsuppsplacesa} and \cref{fig:figsuppsplacesb} show the qualitative comparisons using Places.
\Cref{fig:suppscelebahq} provides the qualitative comparisons using CelebA-HQ.

\figmaskratio

\figdiverserock

\figsuppsplacesA
\figsuppsplacesB

\figsuppscelebahq